%% file: root.tex
\documentclass[letterpaper, 10 pt, conference]{ieeeconf}  

\IEEEoverridecommandlockouts                              

\usepackage[utf8]{inputenc} 
\usepackage[T1]{fontenc}    
\usepackage{hyperref}       
\usepackage{url}            
\usepackage{booktabs}       
\usepackage{amsfonts}       
\usepackage{nicefrac}       
\usepackage{microtype}      
\usepackage{multirow}
\usepackage{subcaption}
\usepackage{wrapfig}
\usepackage{times}
\usepackage{epsfig}
\usepackage{graphicx}
\usepackage{amsmath}
\usepackage{amssymb}
\usepackage{pifont}
\newcommand{\xmark}{\ding{55}}%
\usepackage{booktabs}

\usepackage{lineno}
\modulolinenumbers[5]
\usepackage{graphicx}
\usepackage{booktabs}
\usepackage{amssymb,amsmath,nccmath}
\usepackage{cclicenses}
\usepackage{makecell}
\usepackage{lscape,array}
\newcolumntype{C}[1]{>{\centering\arraybackslash}p{#1}} 
\usepackage[thin, , thinc]{esdiff}
\usepackage{subcaption}
\usepackage{caption}
\usepackage{framed}  
\usepackage{cite}
\usepackage[font=small,skip=0pt]{caption}

\title{\LARGE \bf
Interaction Graphs for Object Importance Estimation in \\
On-road Driving Videos
}

\author{Zehua Zhang$^{1}$*, Ashish Tawari$^{2}$, Sujitha Martin$^{2}$, and David Crandall$^{1}$
\thanks{*Part of this work was done while Zehua Zhang was an intern at Honda Research Institute, USA}
\thanks{$^{1}$The authors are with Indiana University, Bloomington, IN, USA, 47405.
        {\tt\small \{zehzhang, djcran\}@indiana.edu}}%
\thanks{$^{2}$The authors are with Honda Research Institute, San Jose, CA, USA, 95134.
        {\tt\small \{atawari, smartin\}@honda-ri.com}}%
}

\begin{document}

\maketitle
\thispagestyle{empty}
\pagestyle{empty}

\begin{abstract}

A vehicle driving along the road is surrounded by many objects,
but only a small subset of them influence the driver's decisions and
actions. Learning to estimate the importance of each object on the driver's
 real-time decision-making may help better
understand human driving behavior and lead to more reliable
autonomous driving systems. Solving this problem requires models that
understand the interactions between the ego-vehicle and the surrounding 
objects. However, interactions among other objects in the scene
can potentially also be very helpful, \emph{e.g.}, a
pedestrian beginning to cross the road between the ego-vehicle
and the car in front will make the car in front less important. We propose a
novel framework for object importance estimation using an interaction
graph, in which the features of each object node are updated by
interacting with others through graph convolution. Experiments show
that our model outperforms  state-of-the-art baselines with much
less input and  pre-processing.

\end{abstract}

\input{intro.tex}

\input{related.tex}

\input{model.tex}

\input{experiment.tex}

\input{conclusion.tex}

\bibliographystyle{IEEEtran}
\bibliography{IEEEexample}

\end{document}

%% file: intro.tex
\section{INTRODUCTION}
\label{sec:intro}

Driving is a complex task because it involves highly dynamic, complex
environments in which many different autonomous agents (other drivers,
pedestrians, etc.) are acting at the same time.
Human drivers make real-time decisions by combining information from
multiple sources, among which visual information often plays the most
important part. Since humans have foveated vision systems that require
controlling both head pose and eye gaze~\cite{mozer1998computational,
  lazzari2009eye, bowman2009eye, hayhoe2005eye, vidoni2009manual,
  perone2008relation}, people must identify and attend to the most task-relevant objects
in their visual field at any given time.

Learning to predict drivers' attention has become a popular topic in
recent years~\cite{palazzi2018predicting, tawari2017computational,
  tawari2018learning, gao2019goal, xia2018predicting} due to potential
application in advanced driver assistance systems (ADAS) and
autonomous driving. However, much of this
work~\cite{palazzi2018predicting, tawari2017computational,
  tawari2018learning, xia2018predicting} focuses on predicting
pixel-level human eye gaze, which has two main drawbacks. First, drivers
will often look at objects irrelevant to the driving task ---
\emph{e.g.}, beautiful scenery. Second, gaze is limited
to a single, small region at any moment in time, whereas the human may
be actively attending to multiple objects (using short-term memory, peripheral vision,
or multiple saccades) --- for example, 
if a group of people is
crossing the road, a good driver (and hence an autonomous system) should pay attention to all of 
them instead of just a single person.

\begin{figure}[t]
       \includegraphics[width=0.5\textwidth]{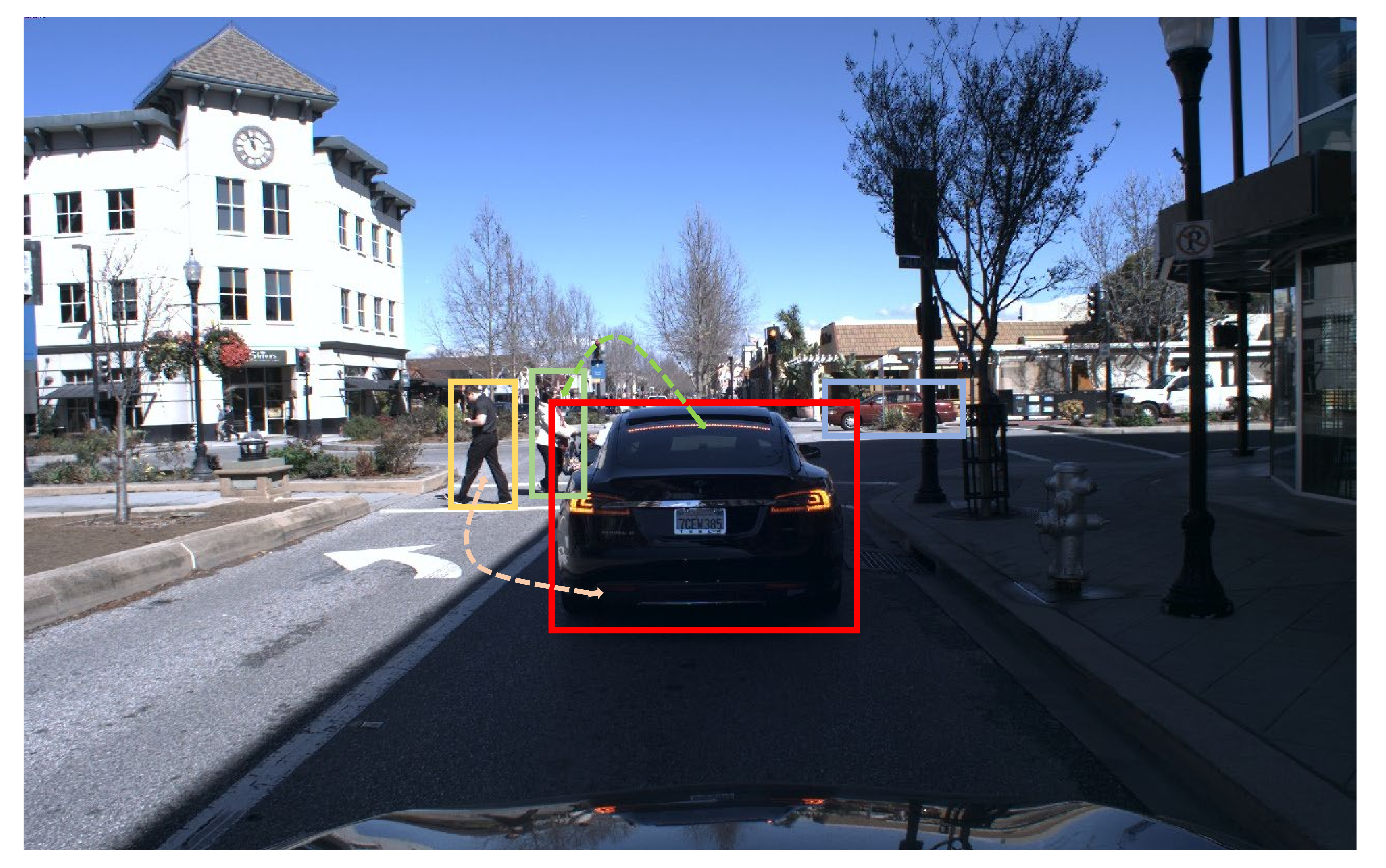}
    \caption{\emph{Given a video clip, our goal is to estimate, in an online fashion,
        which objects in the last frame are important for the driver
        to make real-time control decisions.} We also visualize
      different candidate object boxes, while the red box is the
      ground truth. Without considering the interaction between
      pedestrians and the front car, traditional methods make mistakes
      by predicting both pedestrians and the front car as
      important. Our proposed method can effectively model the
      interactions among other objects (visualized as dashed arrows)
      with a novel interaction graph and thus make the correct
      prediction. In this example, the front car prevents the ego
      vehicle from hitting the pedestrians and thus reduces the
      importance of them.}
    \label{fig:intro}
\end{figure}

To overcome the above problems, in this paper we investigate how to
directly estimate each object's importance to the ego-vehicle for
 making decisions in on-road driving videos without
using eye gaze as an intermediate step, as is shown in
Fig.~\ref{fig:intro}. We use the dataset collected by Gao~\emph{et
  al.}~\cite{gao2019goal}, in which each sample clip was viewed by
experienced drivers and each object in the last frame of the clip was
labeled as either important or not. These on-road videos were recorded
by forward-facing cameras mounted on cars.
This perspective is somewhat different from the
drivers' actual field of view since the dashboard camera is fixed;
it makes the video more stable than if it were from a head-mounted camera,
but also makes the problem more challenging since we cannot use cues
about where the driver was looking
(e.g., human drivers tend to adjust their head pose to center
an attended region within their visual
field~\cite{gteaplusyinli}).

We propose to leverage frequent interactions
among objects in the scene other than the ego-vehicle. Such interactions are often overlooked by other methods but
extremely helpful. For example, in Fig.~\ref{fig:intro}, the front car
will prevent the ego-vehicle from hitting the pedestrians and thus
greatly reduces their importance: the ego-vehicle's driver only needs
to avoid  hitting the car in front at the moment. We model these
interactions using a novel interaction graph, with features pooled
from an I3D-based~\cite{i3d} feature extractor as the nodes, and
interaction scores learned by the network itself as the edges. Through
stacked graph convolutional layers, object nodes interact with each
other and their features are updated from those of the nodes they 
closely interact with. Our experiments show that the interaction
graph greatly improves performance and our model outperforms 
state-of-the-art methods with less input information (RGB clips only) and 
pre-processing (object detection on the target frame only).

%% file: related.tex
\section{Related work}
\subsection{Driver Attention Prediction}

As interest in (semi-)autonomous driving is growing, researchers have
paid more attention to the problem of predicting driver attention. This is typically
posed as predicting pixel-level eye gaze in terms of likelihood
maps~\cite{palazzi2018predicting, tawari2017computational,
  tawari2018learning, xia2018predicting}. Fully convolutional
networks~\cite{long2015fully}, which were original proposed to solve
image segmentation, have been applied by Tawari \emph{et
  al.}~\cite{tawari2017computational,tawari2018learning} for similar
dense spatial probability prediction tasks. Palazzi \emph{et
  al.}~\cite{palazzi2018predicting} combine features from multiple
branches with RGB frames, optical flow, and semantic segmentation to
create the final prediction. Xia~\emph{et al.} propose to handle
critical situations by paying more attention to the frames that are identified as crucial driving moments based on human driver eye gaze movements~\cite{xia2018predicting}.

However, using eye gaze prediction to estimate driver attention has limitations,
such as that drivers can be attending to multiple objects at once (through short-term memory, 
frequent saccades or peripheral vision).
To overcome these drawbacks,
Gao~\emph{et
  al.}~\cite{gao2019goal} collected an on-road driving video dataset
with objects labeled 
as important or not by experienced drivers.  They use an
LSTM~\cite{hochreiter1997long} leveraging goal
information for object importance estimation. To achieve 
state-of-the-art performance, however, their technique requires multiple
sources of input including maneuver information of the planned
path, RGB clips, optical flow, and location information, as well as
complex pre-processing of per-frame object detection
and tracking.

\subsection{Object-level Attention}

While many papers study saliency~\cite{sal1, sal2, sal3, sal4} and
eye gaze~\cite{gaze1, gteaplusyinli, gaze2, gaze3, gaze4, gaze5,
  gaze6, gaze7} prediction, only a few focus on object-level
attention~\cite{obj1,obj2,obj3,obj4,obj5,zhang2019self}. Hand-designed
features are applied for important object and people detection
in Lee \textit{et al.}~\cite{obj1}. Pirsiavash and Ramanan~\cite{obj2} and Ma \textit{et al.}~\cite{obj3} detect objects in hands
as a proxy for attended objects. Bertasius~\emph{et
  al.}~\cite{obj5} propose an unsupervised method for attended object
prediction. The recent work of Zhang et al.~\cite{zhang2019self} tries
to jointly identify and locate the attended object. Inspired by the
foveated nature of human vision system, the class and the location
information are integrated through a self-validation module to further
refine the prediction. But these techniques are for egocentric videos in which both head
movement and ego hands are available for making inferences,
while in our settings the camera is fixed and hand
information is not applicable.

\subsection{Graph Convolutional Networks}

Graph convolutional networks (GCNs)~\cite{kipf2016semi,
  defferrard2016convolutional} have become popular in computer 
vision problems~\cite{gcn1, gcn2, gcn3, yang2018graph} for their
ability to capture long-range relationships in non-Euclidean
space. Compared with traditional convolutional networks which need to stack many layers for a large receptive field, GCNs 
can efficiently 
 model long-range relations with an
adjacency matrix.
 Inspired by these papers, we apply interaction graphs
to solve the problem of object importance estimation in on-road
driving videos. To the best of our knowledge, ours is the first work to
perform per-node predictions in computer vision problems with
GCNs. Also, our model learns to predict the graph edges themselves based on
nodes' interactions, while existing work typically formulates the edges with
hand-designed metrics such as spatial-temporal overlap~\cite{gcn1}.

%% file: model.tex
\begin{figure*}[t]
       \includegraphics[width=1.0\textwidth]{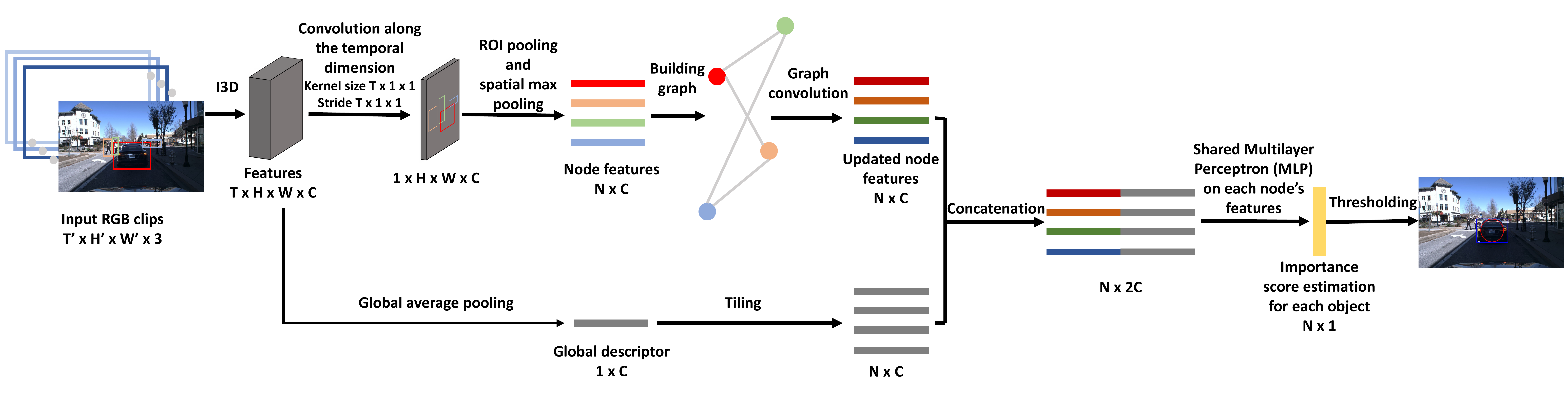}
    \caption{\emph{The architecture of our proposed model.} Features
      from the I3D-based feature extractor are first aggregated
      temporally by convolution along only the temporal
      dimension. Then ROI pooling and spatial max pooling are applied
      to obtain each object node's feature vector. These features are
      updated by interacting with other nodes through graph
      convolution based on the learned interaction graph
      edges. Finally, the updated features are concatenated with the
      global descriptor and fed into a shared Multilayer Perceptron
      (MLP) for object importance estimation. The step of object
      proposal generation is omitted for clarity.}
    \label{fig:model}
\end{figure*}

\section{OUR MODEL}

Given videos taken by forward-facing (egocentric) vehicle cameras,
our aim is to estimate the importance of each object in the
scene to the driver. Following the
same problem setting as~\cite{gao2019goal}, we do online estimation,
predicting object importance scores in the last frame of each 
video clip. We propose a novel framework that leverages the
interactions among objects using graph convolution. An overview of our
model is shown in Fig.~\ref{fig:model}.

\textbf{Object proposals.} Since it is time-sensitive, we would like 
online prediction to 
require as little pre-processing as possible. In contrast to the
state-of-the-art method~\cite{gao2019goal} which requires object
detection on each frame of the input video as well as object tracking,
our model only needs to run object detection on the target frame (the
last frame in the online detection setting). We apply Mask
RCNN~\cite{he2017mask}, using an off-the-shelf
ResXNet-101~\cite{resnet} 32x8d Mask RCNN model~\cite{Detectron2018}
trained on MSCOCO~\cite{lin2014mscoco} without any further
fine-tuning. Proposals of object classes other than person, bicycle,
car, motorcycle, bus, train, truck, traffic light, and stop sign are
removed since they are irrelevant to our task. For simplicity, a number of dummy
boxes with coordinates of $(x1, y1, x2, y2) = (0.07, 0.91, 0.97, 1.0)$
(all box coordinates are rescaled to $[0, 1]$ with respect to the
height and width of the input frame) are padded until each
target frame contains the same number of proposals.
We empirically set
the number of object proposals per target frame after padding to be
$N=40$. The dummy box is the hood of the ego car, for which the
image content almost remains unchanged for all of the samples. Also,
object proposals were pre-generated to save time
in our experiments, though during model design we tried our best to
reduce pre-processing steps to make it meet the time requirements of
online prediction.

\textbf{Visual feature extractor.} Since only RGB clips are used as
input, making correct object importance predictions requires as strong
a feature extractor as possible. We use
Inception-V1~\cite{szegedy2015inception} I3D~\cite{i3d} because of its
capacity for capturing both rich spatial and temporal features.
Temporal motion information is important for reasoning about both the
ego-vehicle's and other objects' intention and future movement, and
spatial appearance information helps determine the inherent
characteristics of each object. Given $T'$ contiguous RGB frames
$B_{t, t + T' - 1} \in R ^ {T' \times H' \times W' \times 3}$, we feed
them through I3D and extract features $F \in R ^ {T \times H \times W
  \times C}$ from the last mixing layer of rgb\_Mixed\_5c. We have $T
= \frac{T'}{8}$, $H = \frac{H'}{32}$, $W = \frac{W'}{32}$, and
$C=1024$ based on the architecture setting of Inception-V1 I3D.

\textbf{Feature vectors of graph nodes.} Extracted features $F$ from
above are further aggregated temporally through one-layer convolution
along only the temporal dimension, with
kernel size and
stride set to $T \times 1 \times 1$. From the obtained feature maps
$F' \in R ^ {1 \times H \times W \times C}$, the features $\{f_i|i =
1,2,..., N\}$ for each object are pooled using ROI
Pooling~\cite{he2017mask, fasterrcnn}. Following~\cite{he2017mask}, we
have $f_i \in R^{7 \times 7 \times C}$ (the temporal dimension is
removed). These feature maps then go through a spatial max pooling
layer, resulting in feature vectors $\{v_i \in R^C|i = 1,2,..., N\}$
for each object node.

\textbf{Graph edge formulation.} The strength of an edge $E_{ij}$ in
our interaction graph should reflect how closely two connected
objects $i$ and $j$ interact with each other. We propose that
the network itself learns to model the edge strength $E_{ij}$ by estimating
how closely two nodes $i$ and $j$ interact with each other.
%
 %
 %
Given node features $v_i$ and $v_j$, an interaction score $IS_{ij}$ is first computed,
    \begin{equation}
        IS_{ij} = \Phi(\Gamma(v_i)||\Gamma'(v_j)), ~~~~~ i,j \in \{1,2,...,N\},
    \end{equation}
    where $\Gamma(\cdot)$ and $\Gamma'(\cdot)$ are linear
    transformations with different learnable parameters $w$ and
    $w'$. $\Phi(x) = \phi x$ is a linear transformation with $\phi$ as
    learnable parameters and $||$ denotes concatenation. With an
    interaction matrix $IS$ obtained by computing interaction scores
    for each pair of nodes, we calculate $E_{ij}$ by applying softmax
    on $IS$ as well as adding an identity matrix $I_N \in R^{N \times
      N}$ to force self attention,
    \begin{equation}
    \label{selfattn}
    E_{ij}=\frac{e^{IS_{ij}}}{\sum^N_{k=1} e^{IS_{ik}}} + I_N, ~~~~~ i,j \in \{1,2,...,N\}.
    \end{equation}
    In this way, the model learns an interaction graph itself based on
    each node's features. The learned edge strength $E_{ij}$ indicates
    how much node $j$ will affect updating node $i$'s features
    through graph convolution and thus reflects how closely they are
    interacting.
    
    We note that the interaction graph learned here is a
    directional graph, as $\Gamma(\cdot)$ and $\Gamma'(\cdot)$ are
    different transformations. This is reasonable since how much node
    $i$ affect node $j$ is not necessarily the same as how much $j$
    affects $i$. For example, in Fig.~\ref{fig:intro}, while the front
    car will greatly reduce the importance of the pedestrians, the
    pedestrians almost have no influence on how important the front
    car is to the ego vehicle.

An alternative way 
of forming the graph is based on feature similarity
between pairs of nodes following~\cite{gcn1}. We ran experiments to
compare our interaction graph with that similarity graph, and found that the
model is not able to learn well with similarity graphs. Our hypothesis is
that similarity graphs are not suitable for our problem, as objects
sharing similar appearance or motion do not necessarily closely
interact, and vice versa. For example, in Fig.\ref{fig:intro}, the
front car and the pedestrians are very different in terms of both
appearance and motion, despite the close interaction between
them. Since our proposed interaction graph yields better
performance, we use it in our experiments here.

\textbf{Graph convolution.} With the graph $E$ formulated, we perform
graph convolution to allow nodes to interact and update each
other. One layer of graph convolution can be represented as:
\begin{equation}
    V' = \sigma(EVW)
\end{equation}
where $E \in R^{N \times N}$ is the edge matrix, $V = [v_1, v_2, ...,
  v_N] \in R^{N \times C}$ is the input node feature matrix, and $W
\in R^{C \in C'}$ is the weight matrix of the layer. $\sigma(\cdot)$
is a non-linear function; we use ReLU in our
experiments. We stack $3$ graph convolutional layers and simply set
$C'=C$ for all of them. After the graph convolution, we obtain an updated
feature matrix $U \in R^{N \times C}$.

\textbf{Per-node object importance estimation.} We now perform
per-node object importance estimation. Although each node's features
are updated through GCN to capture long-range relationships with other
nodes, some global context may still be missing because object
proposals cannot cover the whole image. Also, the object detector is
not perfect and useful objects may be missed (\emph{e.g.,} 
small objects such as traffic lights). To circumvent this problem, we
first apply global average pooling on the extracted features $F$ from
I3D to obtain a global descriptor $D \in R^{1 \times C}$, which are then
tiled $N$ times and concatenated with the updated node features
$U$. Each row of the resulting features $Y \in R^{N \times 2C}$ is fed
into a shared Multilayer Perceptron (MLP) for the final importance
score estimation,
\begin{equation}
    \widehat{score}_i = sigmoid(MLP(Y_i)), ~~~~~ i \in {1,2,...,N}.
\end{equation}
\textbf{Other implementation and training details}
\begin{itemize}
    \item 
    \textbf{Missing detections.}  Since we apply multi-fold cross
    validation (as in~\cite{gao2019goal}) in our experiments, each
    sample will either serve as a training sample or a testing
    sample. For those samples with ground truth objects not detected
    by our object detector, we prepare two sets of proposals for
    training and testing respectively. The set for testing is the same
    as the $40$ proposals obtained by padding dummy boxes to the
    results of Mask RCNN, while the set for training is slightly
    different as we replace dummy boxes with the ground truth object
    boxes which were not detected to avoid misleading the
    network. Note that missing detections of ground truth objects
    still happen when these samples serve as testing data, and thus
    our model can never reach $AP=100\%$.

    \item
    \textbf{Hard negative mining and loss functions.}  The dataset
    collected by Gao~\emph{et al.}~\cite{gao2019goal} suffers from
    significant imbalance: of 8,611 samples (short video clips), only
    4,268 objects are labeled as important. Considering our setting
    of 40 box nodes per sample, the ratio of the total number of
    positive boxes over negative boxes is almost 1:80. In contrast
    to~\cite{gao2019goal} which applies weighted-cross-entropy based
    on the number of positive and negatives boxes, we do hard negative
    mining to address the problem. During each batch, we first compute
    the losses for each node importance estimation $\widehat{score}_i$
    with binary cross entropy loss,
    
    \begin{equation}
    \label{loss1}
        L_{node_i} = L_{be}(score_i, \widehat{score}_i), ~~~~~ i \in {1,2,...,N},
    \end{equation}
    where $score_i$ is the corresponding ground truth and,
    \begin{equation}
    \label{loss2}
        L_{be}(x, \hat{x}) = -xlog(\hat{x}) - (1-x)log(1-\hat{x}).
    \end{equation}
    Then the losses for negative nodes are sorted and we take only the
    $N_{neg}$ greatest from them, along with the losses for all the
    positive nodes, to compute the total loss. Letting $N_{pos}$
    denote the total number of positive nodes, we empirically found
    that $ N_{neg} = max(5\cdot N_{pos}, 10)$ works well (after
    multiple experiments on different ratios).  Supposing $\Psi_{pos}$
    and $\Psi_{neg}$ denote the sets of indices for all the positive
    nodes and the selected negative nodes whose node losses are among
    the top $N_{neg}$, the total loss is then,
    \begin{equation}
    \label{totalloss}
        L_{total} = \frac{1}{N_{pos}}(\sum_{j \in \Psi_{pos}}L_{node_j} + \sum_{k \in \Psi_{neg}} L_{node_k}).
    \end{equation}
    \item
    \textbf{Other details.} We implemented the model with
    Keras~\cite{keras} and Tensorflow~\cite{tensorflow}. A batch
    normalization layer~\cite{batchnorm} was inserted after each layer
    of the feature extractor, with momentum  0.8. We used
    RGB clips with dimension $16 \times 200 \times 320 \times
    3$ as the only input. During training, the I3D feature extractor
    was initialized with weights pretrained on
    Kinetics~\cite{kay2017kinetics} and
    ImageNet~\cite{deng2009imagenet}, while other parts were randomly
    initialized. We trained the model with stochastic gradient descent
    with initial learning rate 0.0003, momentum 0.9, decay 0.0001,
    and L2 regularizer 0.0005. The loss function to be optimized was as
    Eq.~\ref{totalloss}. When making inference, the model  predicts
    an importance score in the range of $[0,1]$ for each of the object
    proposals, resulting in a $40 \times 1$ prediction per sample.
\end{itemize}

%% file: experiment.tex
\section{EXPERIMENTS}
\subsection{Experiment Settings}
\textbf{Dataset.} We evaluate our model on the the on-road driving
video dataset used by Gao \textit{et al.}~\cite{gao2019goal}. This dataset consists of
8,611 annotated samples, each of which is a 30-frame RGB clip (during
our experiments we only use the last 16 of them) recording real-world
driving in highly-dynamic environments. The dataset focuses on real-time
driving around road intersections, and were annotated by
experienced drivers with object importance estimates. 
Please refer to~\cite{gao2019goal}
for detailed
statistics about the dataset.
 In our experiments, we follow the same data
split as in~\cite{gao2019goal} and also perform 3-fold cross validation.

\textbf{Metrics.}  We compute 11-point average precision
(AP)~\cite{pascal-voc-2007} for each data split and then take the
average across the 3 splits. A predicted box is considered as
correct if its IOU with one of the unmatched ground truth boxes is over
0.5. We do not allow duplicated matching to the same ground
truth box. Also, note that due to false negatives of the object
detector, the upper bound performance of our model can never achieve
$100\%$,  as in~\cite{gao2019goal}.

\begin{table*}
\centering
\captionsetup{justification=centering}
      \caption{Comparison of our model with other baselines in terms
        of required input, required pre-processing, and average
        precision on different splits. Our model outperforms the state-of-the-art with the least input and the easiest
        pre-processing.}  {\footnotesize{\textsf{
\begin{center}
\scalebox{0.8}{
      \begin{tabular}{l|*7c|*4c}
\toprule
        Models  & RGB clips & Input length & Optical flows & Goal & Location & Object detection  & Tracking & $AP_1~\uparrow$ & $AP_2~\uparrow$ & $AP_3~\uparrow$ & $avgAP~\uparrow$\\ \midrule
        Our model & \checkmark & 16 & \xmark & \xmark & \xmark & On the target frame & \xmark & 68.5 & \textbf{73.8} & \textbf{71.9} & \textbf{71.4} \\ 
\midrule
        Goal-Visual Model~\cite{gao2019goal}  & \checkmark & 30 & \checkmark & \checkmark & \checkmark & On each frame & \checkmark & \textbf{70.2} & 70.3 & \textbf{72.0} & 70.8 \\ 
        Visual Model~\cite{gao2019goal}  & \checkmark & 30 & \checkmark & \xmark & \checkmark & On each frame & \checkmark & 68.1 & 68.1 & 70.9 & 69.0 \\ 
        Goal-Geometry Model~\cite{gao2019goal}  & \xmark & 0 & \checkmark & \checkmark & \checkmark & N/A & \checkmark & 32.1 & 40.6 & 41.8 & 38.2 \\ 
        Visual Model-Image~\cite{gao2019goal}  & \checkmark & 1 & \xmark & \checkmark & \checkmark & On the target frame & \checkmark & 35.5 & 42.1 & 32.6 & 36.7 \\ 
\bottomrule
      \end{tabular}}
\end{center}
  }}}

      \label{tab:compa}
\end{table*}

\textbf{Baselines.}  We compare our model with the state-of-the-art
goal-oriented model as well as other competitive baselines
in~\cite{gao2019goal}. These baselines have similar network
structures based on LSTMs~\cite{hochreiter1997long}, while the main
difference between them is the input features. The strongest one,
Goal-Visual Model, takes as input an RGB clip of 30 frames along with
optical flow, goal information, and location
information. Tab.~\ref{tab:compa} compares our model with these
baselines from the perspective of the inputs and pre-processing required.
\begin{figure*}[t]
       \includegraphics[width=1.0\textwidth]{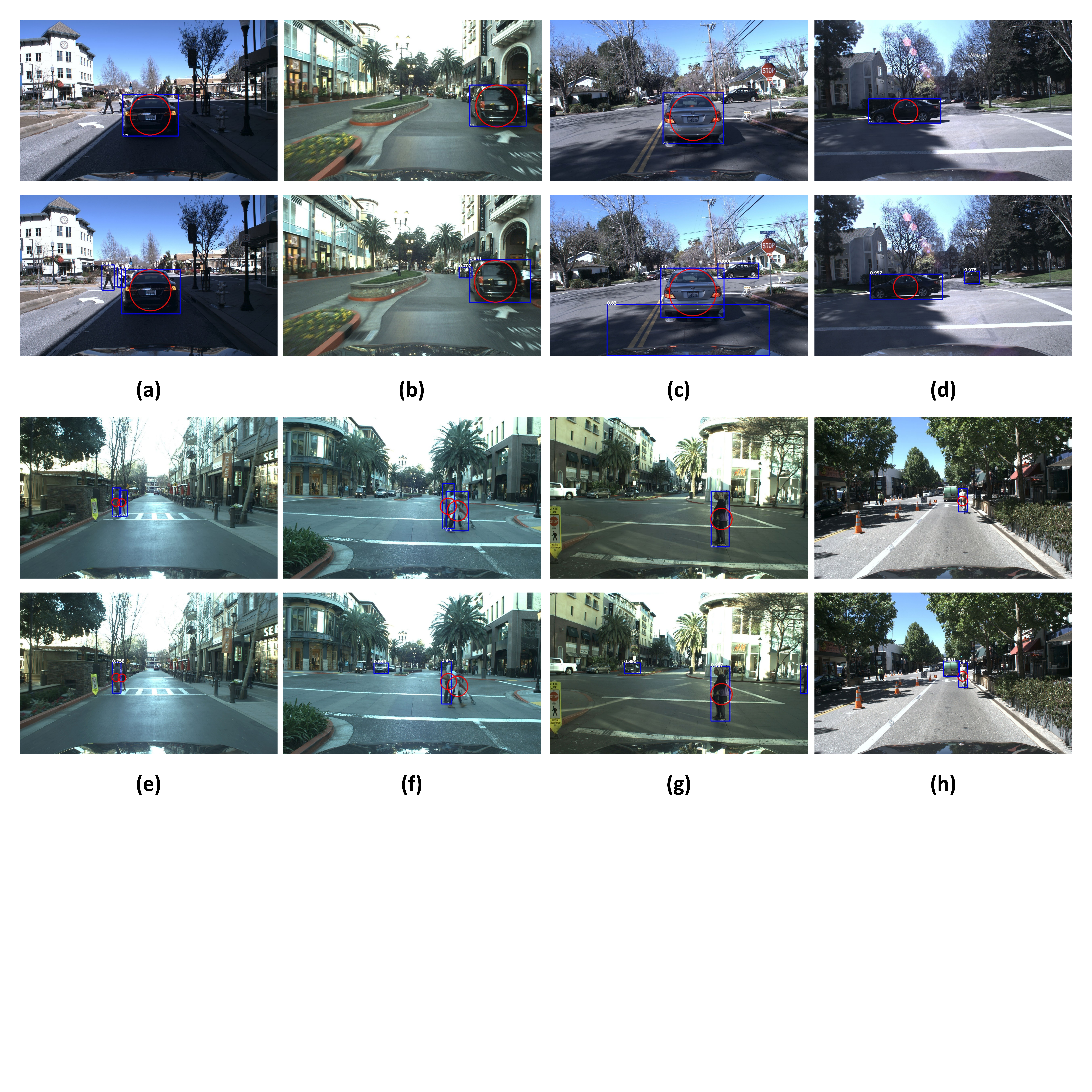}
    \caption{\emph{Qualitative comparison of our model (the 1st and the 3rd rows) with the state-of-the-art method of Goal-Visual Model~\cite{gao2019goal} (the 2nd and the 4th rows).} Blue rectangles are the predicted important object boxes and red circles represent the ground truth. In the upper half (a, b, c and d) we visualize samples with vehicles as the important object, while samples with important pedestrians are visualized in the bottom half. In a,b,c,d,g and h, our model outperforms~\cite{gao2019goal} by making fewer false positive predictions, while in e and f it yields more true positive predictions.}
    \label{fig:compa}
\end{figure*}

\subsection{Experiment Results}
\textbf{Qualitative results.} We qualitatively compare our model with
the Goal-Visual model in Fig.~\ref{fig:compa}. While the Goal-Visual
model usually fails in scenarios with multiple potentially-important
objects (Fig.~\ref{fig:compa}a, b, c, d, g and h), our model 
suppresses false positive predictions significantly by leveraging object
interactions with our proposed interaction graph. Interestingly, when
generating the visualization, our model uses a threshold of 0.3 and
the Goal-Visual model uses 0.5, yet our model still makes fewer false
positive predictions. Our hypothesis is that when multiple objects
can possibly be important, a suppression procedure is performed among
nodes of these objects inside the interaction graph. These nodes
interact with each other and make inferences based on node features to
suppress the false positives. With the interaction graph, our model
becomes more cautious about predicting multiple objects as
important. However, it does not lose the ability to predict multiple
true positives. As shown in Fig.~\ref{fig:compa} (e) and (f) where
the pedestrians are crossing the road together with each other, our
model with an interaction graph can effectively capture the relation
and assign similar importance scores to them.

\textbf{Quantitative results.} Our model and other baselines are
quantitatively compared in Tab.~\ref{tab:compa}. Despite requiring the
least input and the easiest pre-processing, our model still
outperforms the state-of-the-art model in terms of average AP across
the 3 splits. We also observed that our model significantly
outperforms the Goal-Visual model on split 2, achieves comparable AP
on split 3, but performs worse on split 1. The reason may be that the
number of samples on which goal information can significantly help the
final estimation varies across the 3 splits. As we will show in the
next section, in some cases it is almost impossible for the network to make
correct estimates without knowing the goal of the ego-vehicle. In the future,
we plan
to have human annotators further analyze the 3 splits to investigate this.

\begin{figure}[t]
       \includegraphics[width=0.5\textwidth]{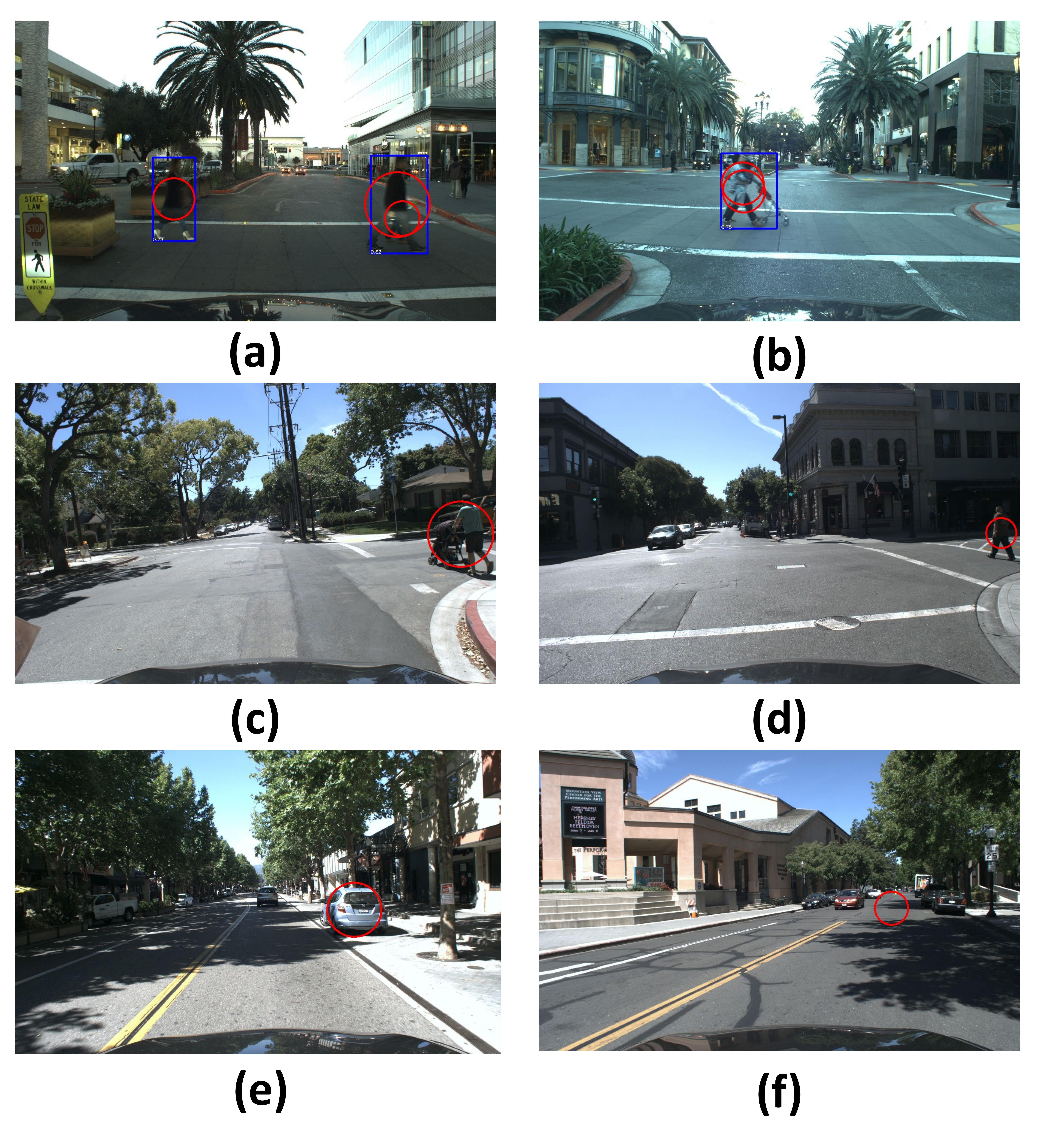}
    \caption{\emph{Sample failure cases of our model.} Blue rectangles are the predicted important object boxes and red circles represent the ground truth. (a) and (b) are caused by  missing detections. (c) and (d) are due to lack of goal information. (e) and (f) are samples with confusing ground truth.}
    \label{fig:fail}
\end{figure}

\textbf{Failure cases.} Sample failure cases are visualized in
Fig.~\ref{fig:fail}. The first rows show failures caused by missing
detections, which  are not the fault of our model as the proposals are
generated by off-the-shelf third-party object detectors. Hopefully in
the future, this kind of failure can be solved with better object
detection models. The failures in the second row reflect the
difficulty of online prediction as the future is unknown. In both of
Fig.~\ref{fig:fail} (c) and (d), the ego-vehicles have been going
straight and stop right at the intersection in the frames. Without goal
information, the model can never know that the driver plans to turn
right, and thus fails to predict the pedestrians crossing the road at
the right as important. To solve this problem requires further
incorporating goal information to our model. The last row shows
samples with confusing ground truth. Fig.~\ref{fig:fail} (e) shows a
common case in which the annotator labeled parked cars along the
road as important even though no one is starting or driving them, while
Fig.~\ref{fig:fail} (f) contains incorrect ground truth of part of the
road region as an important object.

\subsection{Ablation Studies}
We performed ablation studies to evaluate the effectiveness of the proposed
interaction graph, as well as 
some of our
model design choices.

\begin{table}
      \caption{\textit{Results of ablation studies.}}
  {\footnotesize{\textsf{
\begin{center}
\scalebox{0.9}{
      \begin{tabular}{lcccc}
\toprule
        Model  & $AP_1~\uparrow$ & $AP_2~\uparrow$& $AP_3~\uparrow$& $avgAP~\uparrow$\\ 
\midrule
        Our full model & \textbf{68.5} & \textbf{73.8} & \textbf{71.9} & \textbf{71.4} \\ 
\midrule
        Remove interaction graph  & 65.0 & 70.6  & 69.3 & 68.3\\ 
        Remove global descriptor  & 66.0 & 71.7 & 70.3 & 69.3\\ 
        Remove self attention  & 4.2 & 5.3 & 5.8 & 5.1\\ 
        \bottomrule
      \end{tabular}}
\end{center}
  }}}

      \label{tab:ablation}
\end{table}

\textbf{Interaction graph.} We remove the interaction graph along with
the graph convolution layers and directly concatenate the pooled
features of each object proposals with the global descriptor. The
concatenated features are fed into the shared MLP for final
estimation. The results are in Tab.~\ref{tab:ablation}. The AP on each
split drops by $3.5\%$, $3.2\%$, and $2.6\%$, respectively, and the
average AP falls below the score of the Visual model
in~\cite{gao2019goal}, indicating that the interaction graph is
important for performance improvement.

\textbf{Global descriptor.} We remove the global descriptor and let
the model make importance estimation based only on the updated
node features through GCN. Performance drops on all three of the splits, as
shown in Tab.~\ref{tab:ablation}. This implies that the global
descriptor is helpful as it provides useful global context. Also, we
observed that AP drops less by removing the global descriptor than
removing the interaction graph, suggesting that the interaction graph
is more important in improving the performance.

\textbf{Self attention.} The identity matrix $I_N$ in
Eq.~\ref{selfattn} is removed and we trained the model with the new
graph. We found that the model is not able to learn without the forced
self attention, as is shown in Tab.~\ref{tab:ablation}. Self attention
is crucial as it ensures that each node can retain its own
characteristics while interacting with others in graph convolution.

%% file: conclusion.tex
\section{CONCLUSION}
We propose a novel framework for online object importance estimation
in on-road driving videos with interaction graphs. The graph edges are
learned by the network itself based on the nodes' features and reflect how
closely the connected nodes interact with each other. Through graph
convolutional layers, object nodes are able to interact with each
other in the graph and update each other's node features. Experiments
show that our model outperforms the state-of-the-art with much
less input and much easier pre-processing, and ablation studies demonstrate
the effectiveness of the interaction graph as well as
our other model design choices.

\section{Acknowledgments}
Part of this work was done while Zehua Zhang was an intern at Honda Research Institute, USA.
This work was also partially supported 
by the 
National Science
Foundation (CAREER IIS-1253549) and  
by the  Indiana University Office of the Vice
Provost for Research, the College of Arts and Sciences,
and the School of Informatics, Computing, and Engineering
through the Emerging Areas of Research Project \textit{Learning: Brains,
Machines, and Children.}